\documentclass[
]{ceurart}

\usepackage{listings}
\usepackage{graphicx}
\usepackage{multirow}
\graphicspath{ {./images/} }
\usepackage{caption}
\usepackage{subcaption}

\begin{document}

\copyrightyear{2023}
\copyrightclause{Copyright for this paper by its authors.
  Use permitted under Creative Commons License Attribution 4.0
  International (CC BY 4.0).}

\conference{ReNeuIR'23: Workshop on Reaching Efficiency in Neural Information Retrieval,
  July 23--27, 2023, Taipei, Taiwan}  

\title{LACoS-BLOOM: Low-rank Adaptation with \\
           Contrastive objective on 8 bits Siamese-BLOOM}


\author[1]{Wen-Yu Hua}[%
email=wenyu_hua@apple.com
]
\cormark[1]
\address[1]{Seattle, Apple, USA}
 
\author[2]{Brian Williams}[%
email=brian_d_williams@apple.com
]

\author[2]{Davood Shamsi}[%
email=davood@apple.com
]
\address[2]{Cupertino, Apple, USA}

\cortext[1]{Corresponding author.}

\begin{abstract}
Text embeddings are useful features for several NLP applications, such as sentence similarity, text clustering, and semantic search. In this paper, we present a Low-rank Adaptation with a Contrastive objective on top of 8-bit Siamese-BLOOM, a multilingual large language model optimized to produce semantically meaningful word embeddings. The innovation is threefold. First, we cast BLOOM weights to 8-bit values. Second, we fine-tune BLOOM with a scalable adapter (LoRA) and 8-bit Adam optimizer for sentence similarity classification. Third, we apply a Siamese architecture on BLOOM model with a contrastive objective to ease the multi-lingual labeled data scarcity. The experiment results show the quality of learned embeddings from LACoS-BLOOM is proportional to the number of model parameters and the amount of unlabeled training data. 
With the parameter efficient fine-tuning design, we are able to run BLOOM 7.1 billion parameters end-to-end on a single GPU machine with 32GB memory. Compared to previous solution Sentence-BERT, we achieve significant improvement on both English and multi-lingual STS tasks.

\end{abstract}


\maketitle

\begin{figure*}[h]   
  \includegraphics[width=\linewidth]{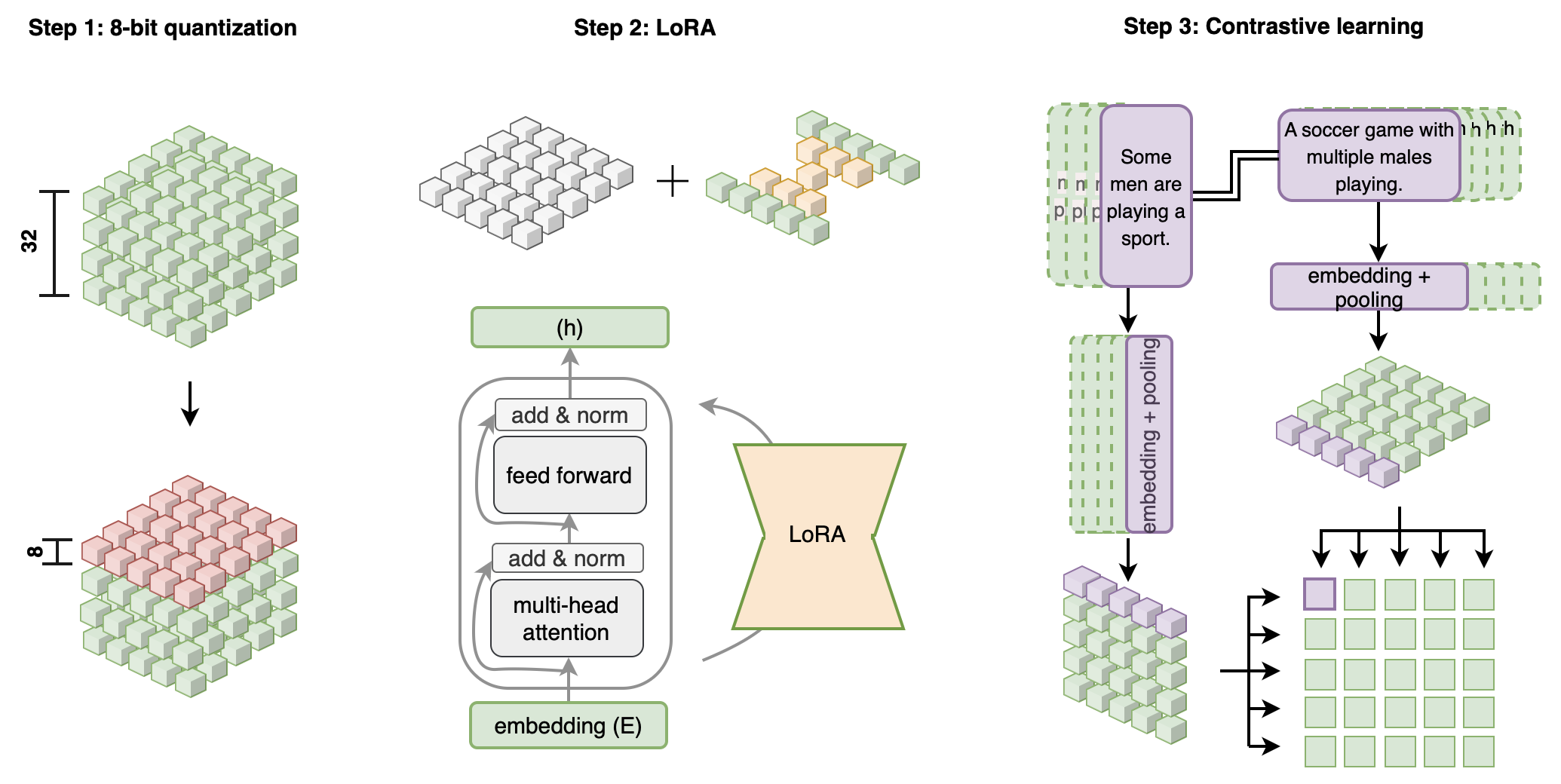}
  \caption{          
LACoS-BLOOM Design. Consider a group of green cubes representing a transformer module. In Step 1, we first quantize the model parameters from 32 float points (green cubes) into 8-bit integers (red cubes). In Step 2, we fine-tune the model by freezing the parameters (gray cubes) and enabling only less than 1\% of the adapters (green and orange cubes), where the orange cubes represent the number of adapters to tune. In Step 3, we only use the entailment class premise and hypothesis pairs from the NLI datasets for training, and we apply a Siamese architecture with MNR contrastive objective to improve performance (the purple cube represents the positive pair, while the others represent negative pairs.)}
  \label{fig:teaser} 
\end{figure*}

\section{Introduction}
Large Language Models (LLMs) are capable of generating human-like language and can be utilized for a wide range of applications, including question answering, summarization, and more. The performance of natural language tasks typically improves as the scale of the model increases \cite{kaplan2020scaling}. Therefore, modern language models have hundreds of billions of parameters \citep{brown2020language, chowdhery2022palm, smith2022using}. Any mention of LLMs is likely to spark discussion around decoder-only Transformer models, where the objective is to predict the next token in a sequence \citep{radford2018improving,radford2019language,brown2020language}. However, the text embedding model is equally important. Text representation, also known as text embedding, is the output of an encoder-based Transformer \cite{devlin2018bert,liu2019roberta}. It is designed to capture the meaning of texts that can be applied to downstream tasks, such as retrieval and clustering. Sentence-BERT \citep{reimers2019sentence} is a classic model for generating similar text representations. It is built on top of BERT \citep{devlin2018bert} and then applied with a Siamese architecture on sentence pairs to classify if a pair is paraphrase identical. As a result, similar context words will have closer embedding representations. Although Sentence-BERT has been successful in several applications in both industry and academia, it only supports English and is a relatively small model.

BigScience Large Open-science Open-access Multilingual Language Model (BLOOM) was released in 2022 \cite{scao2022bloom} and it was trained from 46 natural languages and 13 programming languages. The training datasets cover many research questions surrounding advanced topics such as capabilities, limitations, potential improvements, bias, ethics, environmental impact, and the general AI cognitive research landscape \cite{laurenccon2022bigscience}. To the best of our knowledge (Feb. 2023), BLOOM is the largest publicly available LLM in natural language processing (NLP). The largest BLOOM has 176 billion parameters. BLOOM is powerful, but it is an autoregressive language model aimed at natural language generation. Although it has achieved state-of-the-art (SOTA) performance on several unsupervised NLP tasks, for domain-specific tasks, such as generating semantically meaningful representations, we still need to fine-tune the pre-trained LLM. Our initial attempt is to fine-tune BLOOM with Siamese architecture. However, BLOOM is trained with large-scale parameters on a cluster with hundreds of GPUs, which is less realistic for many situations. In addition, well-performing text embeddings normally require a large amount of labeled data, which is another limitation as the usefulness of multi-lingual labeled data is scarce and expensive.

To overcome the challenges, we propose a parameter efficient fine-tuning solution, i.e., Low-rank Adaptation with a Contrastive objective on top of 8-bit Siamese-BLOOM (LACoS-BLOOM). We take inspiration from the work of bitsandbytes \cite{dettmers2022llm}, where the model weights are frozen in 8-bit format (a model with 7.1 billion parameters is reduced from 20Gb down to 6Gb). We then fine-tune BLOOM with less than 1\% of the parameters using Low-Rank Adaptation (LoRA) \cite{hu2021lora} and update the weights with an efficient 8-bit Adam optimizer \cite{dettmers20218}. Lastly, to make the representation semantically meaningful, we train the model on single-class samples with a multiple negative ranking (MNR) objective on a Siamese architecture \cite{gao2021simcse,neelakantan2022text}. With the design of LACoS, we are able to run various BLOOMs into a single GPU (BLOOM model parameters from 560 million (560m) to 7.1 billion (7b1)). On the evaluation semantic textual similarity (STS) tasks, we achieved significant improvements over the baseline Sentence-BERT.

Next, we present the LACoS-BLOOM model in Section \ref{sec:model}. This is followed by the experimental set up and results in Section \ref{sec:experiment}. The related work is in Section \ref{sec:relatedwork}. Finally we conclude the work and discuss the next step in Section \ref{sec:conclusion}.

\section{Model}
\label{sec:model}
       
LACoS-BLOOM  (Fig. \ref{fig:teaser}) is a text embedding model that generates semantically meaningful representations for multilingual texts. Several techniques have been applied to make LACoS-BLOOM more practical with fewer computational resources and to produce high-quality representations. This includes quantizing the large number of model weights using 8-bit block-wise quantization. The model is fine-tuned using a scalable LoRA and 8-bit Adam optimizer. Finally, the model is enhanced by a Siamese network with a MNR loss.

\subsection{8-bit block-wise quantization}
        
\begin{figure}[h!]
 \centering
        \includegraphics[width=0.9\textwidth]{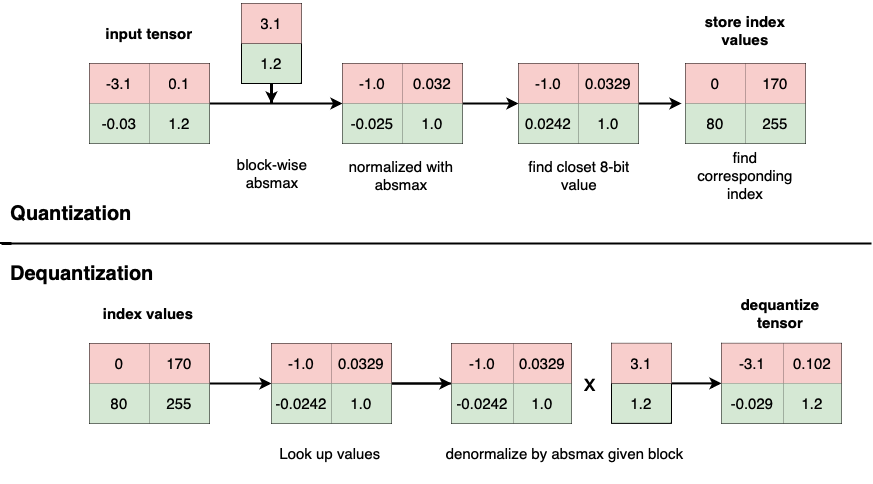}
        \caption{Block-wise quantization and dequantization with block $B=2$ (red and green blocks)}
        \label{fig:8bit}
\end{figure}
We use 8-bit block-wise quantization from \cite{dettmers2022llm}. Figure \ref{fig:8bit} illustrates the steps. A 2 $\times$ 2 matrix split by block size $B=2$ (red and green blocks). Within each block, we find the absolute maximum values, and then use these values to map the float32 weights to 8-bit integer values/indices. Once the weights have been quantized, the indices are stored, which can significantly reduce the footprint of the model. When we update the model, those parameters are de-quantized back to 32 or 16 float points for just-in-time multiplication. The method we use differs from other 8-bit approaches in that we use 8-bit quantization only for storage and perform computations in float16 or float32. This allows the use of nonlinear quantization that is tailored to the distribution of each individual weight, which can reduce error without affecting inference performance.

\subsection{Low-Rank adaptation}
\label{sec:lora}
The adapter approach utilizes small, trainable matrices with low rank to approximate weight updates for downstream tasks. An approach, LoRA, represents updates using a low-rank decomposition in Eq. (\ref{eq:lora}) of the pre-trained weight matrics:
\begin{equation}
        \mathbf{W} + \Delta W = \mathbf{W} + \mathbf{W}_{\text{down}} \times \mathbf{W}_{\text{up}}.
        \label{eq:lora}
\end{equation}
 
The decomposition in Eq. (\ref{eq:lora}) is represented by two tunable matrices, $\mathbf{W}_{\text{down}} \in \mathbb{R}^{d \times r}$ and $\mathbf{W}_{\text{up}} \in \mathbb{R}^{r \times k}$, ($r \ll \text{min}(d,k)$), and is applied to query and value projection matrices in multi-head attention layers from \cite{hu2021lora}. In this work, we apply LoRA to feed-forward linear layers and the last hidden embedding layer, as suggested by previous research \cite{he2021towards}.

\subsection{Siamese network}
\label{sec:siamese}
        
The use of LoRA on LLMs has been successful for fine-tuning domain-specific tasks, but another limitation for finetuning tasks is the availability of labeled data. To address this issue, we propose using a Siamese architecture with contrastive objective, i.e., MNR loss \cite{henderson2017efficient}.
        
MNR loss with Siamese architecture is an approach that allows the model to learn accurate semantic similarity representations despite the limitations of limited labeled data. Given a sequence of mini-batch size $n$, $P={(u_1, v_1), (u_2, v_2),..., (u_n, v_n)}$, where $(u_i, v_i)$ is a positive pair, and $(u_i, v_j)$ for $i \neq j$ are negative pairs. Sentence pairs are passed through LoRA 8-bit BLOOM to obtain the last hidden layer embedding for each token. A mean pooling layer is then applied to produce sentence-level embedding. The similarity score between the embedding pair $(u,v)$ is computed by a cosine function and denoted as $sim(u,v)$. Note that given each mini-batch, there is only 1 positive pair, and others are negatives (denoted $\bar{P}$) (Step 3 in Fig. \ref{fig:teaser}). The goal is to minimize the negative log-likelihood for softmax-normalized scores in Eq (\ref{eq:mnrl}):
        
\begin{equation}
        \mathit{L} = \sum_{(u,v) \in P} log \frac{exp(sim(u,v))}{exp(sim(u,v)) + \sum_{w \in \bar{p}}exp(sim(u,w)}.
        \label{eq:mnrl}
\end{equation}

\begin{figure}[h]
\centering 
        \includegraphics[width=0.80\textwidth]{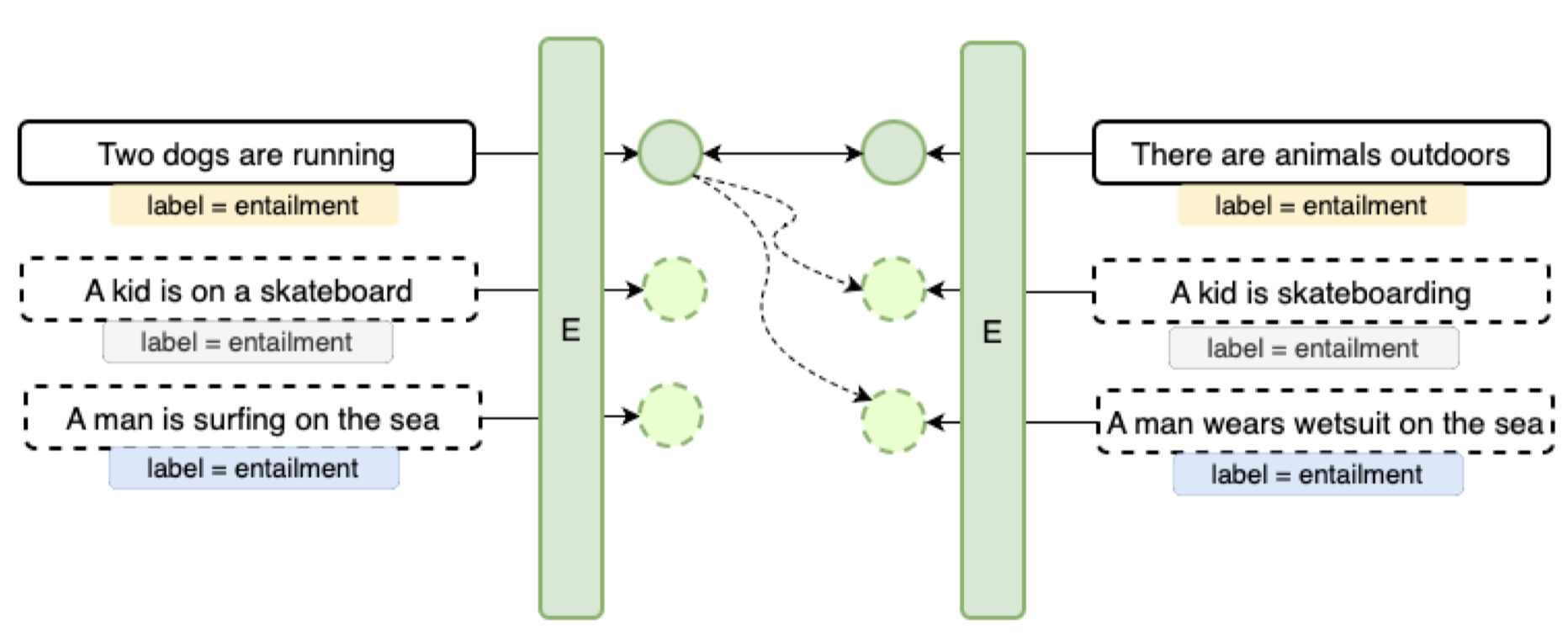}
        \caption{Siamese network with only entailment pair of samples in NLI data; where the solid line shows a positive pair and dashed lines show negative samples given a mini-batch}
        \label{fig:siamese}
\end{figure}

\section{Experiment}
\label{sec:experiment}
\subsection{Experimental setup}
We perform two experiments to train the LACoS-BLOOM model. The first experiment is the Stanford Natural Language Inference (SNLI) \cite{bowman2015large} and Multi-Genre NLI (MNLI) \cite{williams2017broad} datasets, while the second is the Multilingual NLI (multi-NLI) dataset \cite{conneau2018xnli}. During training, we only employ data pairs belonging to the entailment class and apply a Siamese network with MNR loss. Figure \ref{fig:siamese} demonstrates how we created positive and negative samples for every mini-batch. We conduct a grid search for a mini-batch size, choosing between 32 and 64, and the 8-bit Adam optimizer with a learning rate of 1e-4, 2e-5, or 5e-5. We fine-tune the model for 1 epoch. 

We carry out experiments using the LACoS-BLOOM model with sizes ranging from 560m to 7b1 and adapter dimensions ($r$) of 1, 2, 4, 8, or 16. For each BLOOM model size, we retain the best checkpoint for the final evaluation. We utilize the same configuration as Sentence-BERT (SBERT) \cite{reimers2019sentence} with the softmax objective for the baseline, where the pre-trained model is "bert-base-multilingual-cased." The experiments were executed on a single GPU with Volta architecture and 32GB of memory.
       
To select the optimal model, we utilize the test dataset from SNLI and MNLI as our validation set. We aggregate the validation loss and standardize it to a common range. Figure \ref{fig:model_size} displays the number of adapters and the validation error at different BLOOM. The BLOOM 560m demonstrates the lowest validation error with four adapters for each module, whereas the BLOOM 7b1 exhibits the lowest validation error with one adapter per layer. This highlights that when the model size is small, it is necessary to enable more adapters, whereas when the model size is large, only a few adapters are sufficient. 
       
\subsection{Performance comparison}
 
We assess the performance of the LACoS-BLOOM model on seven STS English tasks (STS12-16 \cite{conneau2017supervised}, STS-B \cite{cer2017semeval} and SICK-R \cite{marelli2014sick}) and one multi-lingual STS task (xSTS \cite{stsb_multi_mt}), all of which had not been included in the training process. The STS datasets provide labels ranging from 0 to 5, indicating the semantic relatedness of sentence pairs. We use the same evaluation metric as the baseline SBERT, which is the maximum Spearman's rank correlation among the cosine similarity, Manhattan-distance, Euclidean-distance, and dot-product similarity of sentence embeddings and the golden label datasets. 

\begin{figure}[ht]
\centering 
        \includegraphics[width=0.70\textwidth]{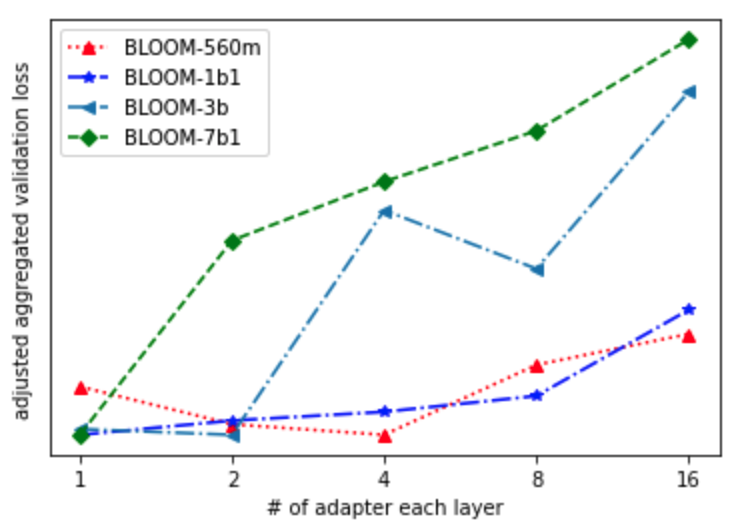}
        \caption{aggregated validation loss vs \# of adapter for each layer}
        \label{fig:model_size}
\end{figure}
         
To evaluate the performance, we use two more metrics: STS-Avg. and xSTS-Avg. STS-Avg. is the average score among STS12-16, STS-B, and SICK-R, which are common benchmarks for evaluating the performance of semantic text similarity models. xSTS-Avg. is the average score across all languages and was used to assess the cross-lingual performance of our model.

We evaluate the BLOOM models identified from Fig \ref{fig:model_size} on the STS tasks and present the correlation scores in Table \ref{tbl:eval}. The scores increased as the model size increased, with LACoS-BLOOM 7b1 achieving the best performance. We use LACoS-BLOOM 7b1 to evaluate the English STS tasks and apply it to xSTS task. Our LACoS-BLOOM method improved the performance of both the English and multi-lingual STS task by at least 4+\%. One observation is that SBERT's performance on multilingual tasks is not as good as it is on English tasks. This could be attributed to the fact that SBERT is a relatively small model, which may limit its ability to transfer knowledge from training to evaluation tasks that differ significantly. 
 
\begin{table}[ht]
\caption{Sentence embedding performance on STS tasks; max(Spearman’s rank correlation) (\%); STS-Avg. is the average score among STS12-16, STS-B and SICK-R. xSTS-Avg. is the average score across all languages}
\label{tbl:eval}
\setlength\tabcolsep{1.7pt}
\begin{tabular*}{\linewidth}{l l c c c c c c c c c c}
\toprule
training & \multicolumn{1}{r}{dataset} & STS12 & STS13 & STS14 & STS15 & STS16 & STS-B &SICK-R & STS-Avg.& xSTS-Avg. \\
data & \multicolumn{1}{l}{model size}\\
\midrule
 \multirow{5}{*} & SBERT& 59.46 & 56.49 &  53.58 & 61.15 & 57.42 & 57.69 & 61.94  & 58.42   & 52.30 \\
\cline{3-11}
SNLI & LACoS-BLOOM-560m & 47.54 &  39.54 & 34.56 & 49.71 & 42.10 & 45.84 & 46.09  & 43.62  & 42.15\\
\cline{3-11}
$+$ & LACoS-BLOOM-1b1 & 59.25 &  62.22 & 56.04 & 62.88  & 59.38 & 59.13  & 54.25  & 59.02 &54.20 \\
\cline{3-11}
MNLI & LACoS-BLOOM-3b1 & 59.74 & 63.81  & 57.12  & 63.95  & 61.78 &61.18 & 55.93 & 60.50  & 56.75 \\
\cline{3-11}
 & LACoS-BLOOM-7b1 & 63.92 & 65.04 & 58.46  & 66.47 & 63.46 & 62.23  & 57.09  & 62.38  & 56.81\\
\midrule
multi- & SBERT & 43.58 &  51.19 &  45.82 & 51.97 & 58.16 & 41.35 & 46.49  & 48.37&  48.82 \\
\cline{3-11}
 NLI & LACoS-BLOOM-7b1 & 66.27 &  71.93 &  67.41 &  75.78 & 71.44 & 70.67  & 57.76  &  68.75  & 70.82 \\
\bottomrule
\end{tabular*}
\end{table}

\subsection{Ablation study}

As BLOOM is an LLM, one advantage is its feasibility for transfer learning. Therefore, we perform zero-shot inference on BLOOM. On the other hand, the previous solution (i.e., simCSE) \cite{gao2021simcse} showed that fine-tuning a full-size model with an MNR contrastive objective achieved the SOTA result on STS tasks. To make a fair comparison, we chose the BLOOM model with 1.1 billion parameters for zero-shot inference, LACoS-BLOOM, and full model fine-tuning, since this is the largest model size that can fit in a single GPU under simCSE setting.

The training data includes SNLI and MNLI for English tasks, and we evaluate the performance on STS tasks. The results are reported in Table \ref{tbl:ablation}. From the results, we find that LACoS-BLOOM outperforms the zero-shot solution. The performance of LACoS-BLOOM is comparable to the full model fine-tuning performance and the computational cost is much more lower.

\begin{table}[ht]
\caption{Ablation study, we trained the LACoS-BLOOM and full size model on SNLI and MNLI datasets, sentence embeddings were evaluated on STS tasks; max(Spearman’s rank correlation) (\%); STS-Avg. is the average score among STS12-16, STS-B and SICK-R. xSTS-Avg. is the average score across all languages}
\label{tbl:ablation}
\setlength\tabcolsep{3.5pt}
\begin{tabular*}{\linewidth}{l l c c c c c c c c c }
\toprule
training & \multicolumn{1}{r}{dataset} & STS12 & STS13 & STS14 & STS15 & STS16 & STS-B &SICK-R & STS-Avg.& \\
data & \multicolumn{1}{l}{model setting}\\
\midrule
SNLI & zero-shot learning & 49.21 & 48.89  & 40.16 & 56.24 & 52.13 & 39.63 &  53.09  & 48.48 \\
\cline{3-11}
$+$ & LACoS-BLOOM-1b1 & 59.25 &  62.22 & 56.04 & 62.88  & 59.38 & 59.13  & 54.25  & 59.02  \\
\cline{3-11}
MNLI & full size model finetune &  69.71 &  73.14 &  68.64 & 76.94 & 76.17 & 70.44 & 68.30  & 71.90\\ 
\bottomrule
\end{tabular*}
\end{table}

\section{Related work}
\label{sec:relatedwork}
\subsection{Compressing LLM}

Deep learning models, particularly transformer-based language models, have achieved SOTA results in NLP, computer vision, speech analysis and other tasks. However, these models can be computationally expensive, so various model compression methods, such as pruning, quantization, knowledge distillation, parameter sharing, tensor decomposition, and sub-quadratic Transformer-based methods, have been developed to reduce computational costs while maintaining performance \cite{ganesh2021compressing,gupta2022compression}. 8-bit quantization is a popular approach for optimization as it reduces both memory and computing requirements without the need to manipulate the architecture and can be used with machine learning frameworks and hardware toolchains. Different from previous work in applying quantization to their application, we use blockwise quantization and dequantization to save the footprint and maintain the perplexity score. As a result, such solution optimizes not just the network but the entire application (e.g., network bandwidth, inference latency and power consumption).

\subsection{Parameter-efficient fine-tuning methods}
In NLP, fine-tuning large pre-trained language models on downstream tasks is common practice but can be impractical as the model size and number of tasks increases. To address this, various parameter-efficient transfer learning approaches have been proposed such as adapters \cite{houlsby2019parameter}, prefix-tuning \cite{li2021prefix} and LoRA \cite{hu2021lora}. The idea behind adapters is to fine-tune large models by only enabling a small subset of parameters on each transformer layer. Prefix-tuning keeps the model parameters frozen and only prepends a few examples to the task input while optimizing the objective based on the controllable prefix texts. LoRA utilizes the adapter approach, but instead of adding a subset of parameters, it enables a few low-intrinsic adapters in parallel with the attention module which does not increase inference latency. In this work, we are interested in LoRA because the design allows for flexibility in adding adapters anywhere \cite{he2021towards}, making it useful for scaling up to large language models for improved performance on specific tasks. 

\subsection{Siamese network with contrastive loss}
Siamese architecture has been widely used in computer vision, NLP, and more. The goal is to learn a similarity function between two inputs. Recent work has shown that the Siamese architecture can boost performance with a self-learning objective in several natural language tasks (e.g., \cite{reimers2019sentence,gao2021simcse,xiong2020approximate}). In this work, we propose to incorporate MNR objective. One advantage of the MNR is that it doesn't rely on labeled data and only considers ranking a set of similar items higher than a set of multiple dissimilar examples. This makes it more efficient in terms of computation and memory, and can lead to a more robust model that generalizes well to new examples.
 
\section{Conclusion and future work}
\label{sec:conclusion}
In this paper, we propose a parameter efficient fine-tuning method called LACoS-BLOOM for extracting multilingual text embeddings from a Large Language Model (LLM). We use 8-bit quantization to reduce the model footprint. We then improve the performance of LLM fine-tuning using LoRA, and further enhance semantic similarity using a Siamese network with MNR. Our solution can train 7.1 billion BLOOM end-to-end on a single GPU. On STS tasks, our method significantly outperforms the baseline as well as zero-shot LLM BLOOM. Our solution is able to scale up the LLM to 7.1 billion model. In the future, we plan to incorporate DeepSpeed with LACoS-BLOOM to efficiently scale up the training task to the full BLOOM.

\bibliography{sbloom}

\end{document}